\documentclass[letterpaper, 10 pt, conference]{ieeeconf}  
\IEEEoverridecommandlockouts
\usepackage{graphicx}
\usepackage{mathptmx} % assumes new font selection scheme installed
\usepackage{times} % assumes new font selection scheme installed
\usepackage{amsmath} % assumes amsmath package installed
\usepackage{amssymb}  % assumes amsmath package installed
\usepackage{units}
\usepackage{verbatim}  % multiline comments via \begin{comment}\end{comment}
\usepackage{nicefrac}		% to get better units
\usepackage{acronym}		% expand first occurance of acronyms
\usepackage{glossaries}		% acronyms + glossaries
\usepackage[usenames,dvipsnames]{xcolor}			% extened color package
\usepackage{cite} % smarter citations, e.g. makes [1][2][3] into [1]-[3]
\usepackage{url} % \url{my_url_here}
\usepackage[multidot]{grffile} % allow dots in graphics filenames
\usepackage{multirow}
\usepackage{booktabs}
\makeatletter
\let\NAT@parse\undefined
\makeatother
\usepackage{hyperref}
\hypersetup{colorlinks,allcolors=black}
\newacronym{ACFR}{ACFR}{Australian Centre for Field Robotics}
\newacronym{USyd}{USyd}{the University of Sydney}
\newacronym{AUV}{AUV}{autonomous underwater vehicle}
\newacronym{UAV}{UAV}{unmanned aerial vehicle}
\newacronym{SLAM}{SLAM}{simultaneous localisation and mapping}
\newacronym{SfM}{SfM}{structure-from-motion}
\newacronym{SNR}{SNR}{signal-to-noise ratio}
\newacronym{DFT}{DFT}{discrete Fourier transform}
\newacronym{FFT}{FFT}{fast Fourier transform}
\newacronym{SIFT}{SIFT}{scale invariant feature transform}
\newacronym{TP}{TP}{true positive}
\newacronym{FP}{FP}{false positive}
\newacronym{BuFF}{BuFF}{burst feature finder}
\newacronym{NIR}{NIR}{near-infrared}
\newacronym{SURF}{SURF}{speeded up robust features}
\newacronym{DoG}{DoG}{difference of Gaussians}
\newacronym{LiFF}{LiFF}{light field features}
\newacronym{ROC}{ROC}{receiver operating characteristic curve}
\newacronym{EV}{EV}{exposure value}

\newacronym{DOF}{DOF}{degree-of-freedom}

\title{\LARGE \bf
BuFF: Burst Feature Finder for Light-Constrained 3D Reconstruction
}

\author{Ahalya Ravendran, Mitch Bryson, Donald G. Dansereau
\thanks{The authors are with the Australian Centre for Field Robotics (ACFR), School of Aerospace, Mechanical and Mechatronic Engineering, The University of Sydney and with the Sydney Institute for Robotics and Intelligent Systems, 2006 NSW, Australia.
{\tt\small ahalya.ravendran, mitch.bryson, donald.dansereau@sydney.edu.au}}%
}

\newcommand{\myquat}[1]{\bar{#1}}

\newcommand{\q}{\myquat{q}}

\newcommand{\Cq}[2]{\boldsymbol{C}(\q)}

\newcommand{\Table}[1]{Tab. #1}

\begin{document}
\maketitle
\thispagestyle{empty}
\pagestyle{empty}

%%%%%%%%%%%%%%%%%%%%%%%%%%%%%%%%%%%%%%%%%%%%%%%%%%%%%%%%%%%%%%%%%%%%%%%%%%%%%%%%
\begin{abstract}
Robots operating at night using conventional vision cameras face significant challenges in reconstruction due to noise-limited images. Previous work has demonstrated that burst-imaging techniques can be used to partially overcome this issue. In this paper, we develop a novel feature detector that operates directly on image bursts that enhances vision-based reconstruction under extremely low-light conditions. Our approach finds keypoints with well-defined scale and apparent motion within each burst by jointly searching in a multi-scale and multi-motion space. Because we describe these features at a stage where the images have higher signal-to-noise ratio, the detected features are more accurate than the state-of-the-art on conventional noisy images and burst-merged images and exhibit high precision, recall, and matching performance. We show improved feature performance and camera pose estimates and demonstrate improved structure-from-motion performance using our feature detector in challenging light-constrained scenes. Our feature finder provides a significant step towards robots operating in low-light scenarios and applications including night-time operations.
\end{abstract}
%%%%%%%%%%%%%%%%%%%%%%%%%%%%%%%%%%%%%%%%%%%%
\section{Introduction}
\label{sec:intro}

%%%%%%%%%%%%%%%% Introduction Chapter %%%%%%%%%%
Robots that use vision sensors for reconstruction face significant challenges in low light environments, as traditional/existing state-of-the-art methods in ~\gls{SLAM} and~\gls{SfM} fail to perform correctly with the resulting noise-limited images. Robots may address this issue by carrying their own light source into the environment, but in many applications this is not possible, for example when studying the behaviour of nocturnal animals (which are light sensitive) or due to weight/power limitations of smaller platforms. There exists a need for image processing techniques that can work effectively with images captured in low light.

Burst imaging is an established mobile photography technique of capturing a scene with high overlap between frames over a small exposure time to produce a single image with an improved~\gls{SNR} upon merging~\cite{Hasinoff2016,Liba2019,Wronski2019,Liu2014}. Burst-merged images demonstrate higher~\gls{SNR} which is favourable for vision-based reconstruction, without the need for additional ambient light sources~\cite{Hasinoff2016}. Adapting burst imaging for robotics, burst-based~\gls{SfM}~\cite{aR2021} established the viability of using burst imaging for reconstruction under extremely low light.

Prior works and on-going developments on burst imaging are steered toward generating all the pixels of a temporally merged image from a burst. Detecting features from such images tends to produce errors due to pairwise misalignment and quantization noise. In this paper, we proposed to address these issues by finding features directly in the burst.

We describe an imaging pipeline that captures multiple bursts along the trajectory of a robot, and find spatio-temporal features within each burst by jointly searching in scale-motion space. We describe these features based on their histograms of edge orientations. This descriptor operates on motion-filtered images, yielding an~\gls{SNR} improvement.

%%%%%%%%%%%%%%%% Abstract Image %%%%%%%%%%
\begin{figure}[t]
	\centering
	\includegraphics[width=\columnwidth]{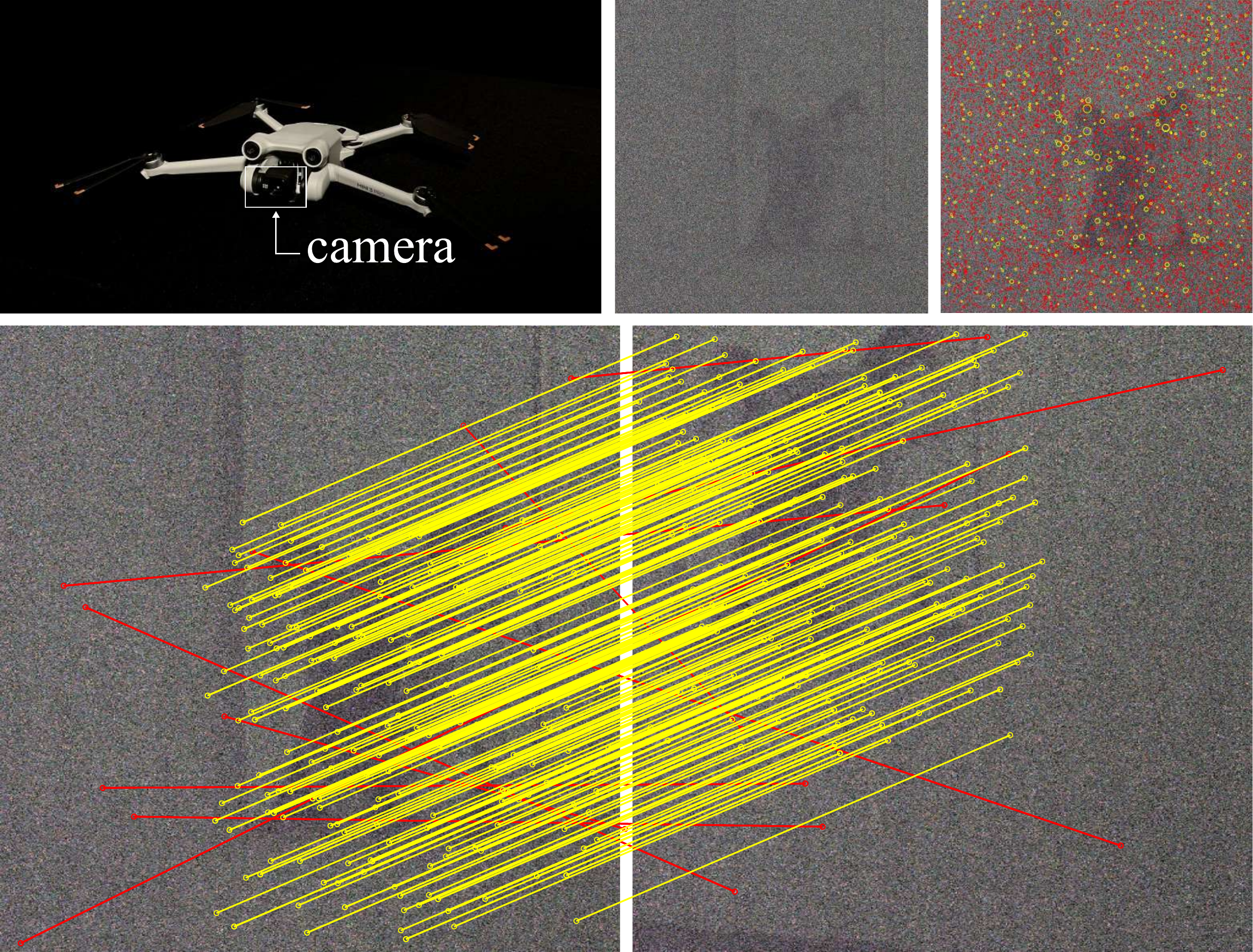}
	\caption{Feature matching in low light: a commercial drone (top-left) captures imagery that is too noisy for conventional 3D reconstruction in low light (top-middle). This is because of the high rate of spurious feature detection and low-quality feature matches offered by conventional features like SIFT (top-right and bottom, red). The proposed BuFF feature yields fewer higher-quality features resulting in many more correctly matched pairs (yellow). In this work we show the BuFF feature enables 3D reconstruction in previously prohibitive low light conditions.}\label{fig_abstract}
\end{figure} % 10 vs 274

Our key contributions are:
\begin{itemize}
\item We introduce~\gls{BuFF}, a 2D + time feature detector and descriptor that finds features with well defined scale and apparent motion within a burst of frames,
\item We propose the approximation of apparent feature motion as either 1D or 2D linear segments under typical robotic platform dynamics, enabling critical refinements relative to prior work on hand-held imagery, and
\item We establish variations of BuFF matched to these apparent motion types and demonstrate it significantly outperforming 2D feature detectors applied to conventional and burst imagery in low-SNR scenes.
%\item We propose two variants of the \gls{BuFF}, one that reduces the computational complexity of feature extraction under certain motion assumptions, and
%\item We make one of our experimental datasets containing low light images and ground-truth images captured in good lighting conditions publically available to assist in further research into low light robotic vision.
\end{itemize}

We validate our method using real burst imagery collected by a robotic arm. Code and dataset are available at \url{https://roboticimaging.org/Projects/BuFF/}.

To evaluate, we compare our results against alternative approaches of using conventional noisy images and burst-merged images for reconstruction, both that uses~\gls{SIFT}~\cite{Lowe2004} for feature extraction. We show our proposed feature extractor finds more true features and fewer spurious features in light-constrained scenes as shown in Fig.~\ref{fig_abstract}. We evaluate the potential of our method to improve 3D reconstruction using the widely available COLMAP SfM software~\cite{Schonberger2016}. With this, we demonstrate our method yielding more complete 3D models containing more 3D points, more accurate camera pose estimates, and higher performance in terms of feature metrics like match score, match ratio and precision comparing to alternative approaches. Importantly, we show COLMAP successfully converging for scenes where it previously could not due to insufficient~\gls{SNR}. 
\section{Related Work}
\label{sec:related}

%%%%%%%%%%%%%%%% Related Chapter %%%%%%%%%%
Feature-based 3D reconstruction from images relies on establishing accurate and reliable pixel-level correspondences between points across multiple images. In robotics, we commonly use~\gls{SIFT} for 2D feature extraction to perform feature-based 3D reconstruction~\cite{Schonberger2016}. \gls{SIFT} builds a~\gls{DoG} pyramid by convolving a single image with a range of scales of {DoG} filters. It detects distinctive keypoints by finding the extrema in the local neighbourhood of~\gls{DoG} pyramid and computes a local descriptor from the normalized region around each detected keypoint based on their histogram of gradients. To find correspondences across other images, the similarity between descriptors is measured. Other 2D feature extractors like~\gls{SURF}~\cite{Bay2006} and ORB~\cite{Rublee2011} also have shown promising results in reconstruction applications including object recognition~\cite{Bay2008} and~\gls{SLAM}~\cite{Mur2017}. The performance of all of these feature detectors reduce in low light.

Another approach to finding features in low light is via machine learning methods~\cite{Dusmanu2019}. These methods require large training dataset and training time. Other studies on low-light reconstruction, use complementary sensors~\cite{Zhuang2021,Chen2022} and expensive cameras~\cite{Shin2019} to reconstruct light-constrained scenes while we demonstrate our method using a conventional camera.
    
Burst imaging is a computation method that captures multiple images with small motion between frames, aligns each frame to a common image within a burst and temporally merges all the aligned images to get a higher~\gls{SNR} image~\cite{Hasinoff2016}. Limitations of this approach have been recently addressed using learning-based methods to an extent~\cite{Mildenhall2018,Luo2022}. In~\cite{aR2021}, the authors adapt burst photography methods for use in improving 3D reconstruction accuracy in low-light scenes.

We propose to take this a step further, by finding features directly on a burst via joint spatio-temporal search instead of finding 2D features on a temporally-merged image. Drawing inspiration from~\gls{LiFF}~\cite{Dansereau2019} and~\gls{SIFT}~\cite{Lowe2004}, in this paper, we develop a feature detector that finds extrema not only in scale-space, but also in the space of apparent motion across an image stack. This allows us to find robust spatio-temporal features in lower \gls{SNR} images for light-constrained reconstruction, by leveraging the motion information between frames in a burst during feature detection. 

\begin{figure}[t!]
\centering
	\includegraphics[width=0.95\columnwidth]{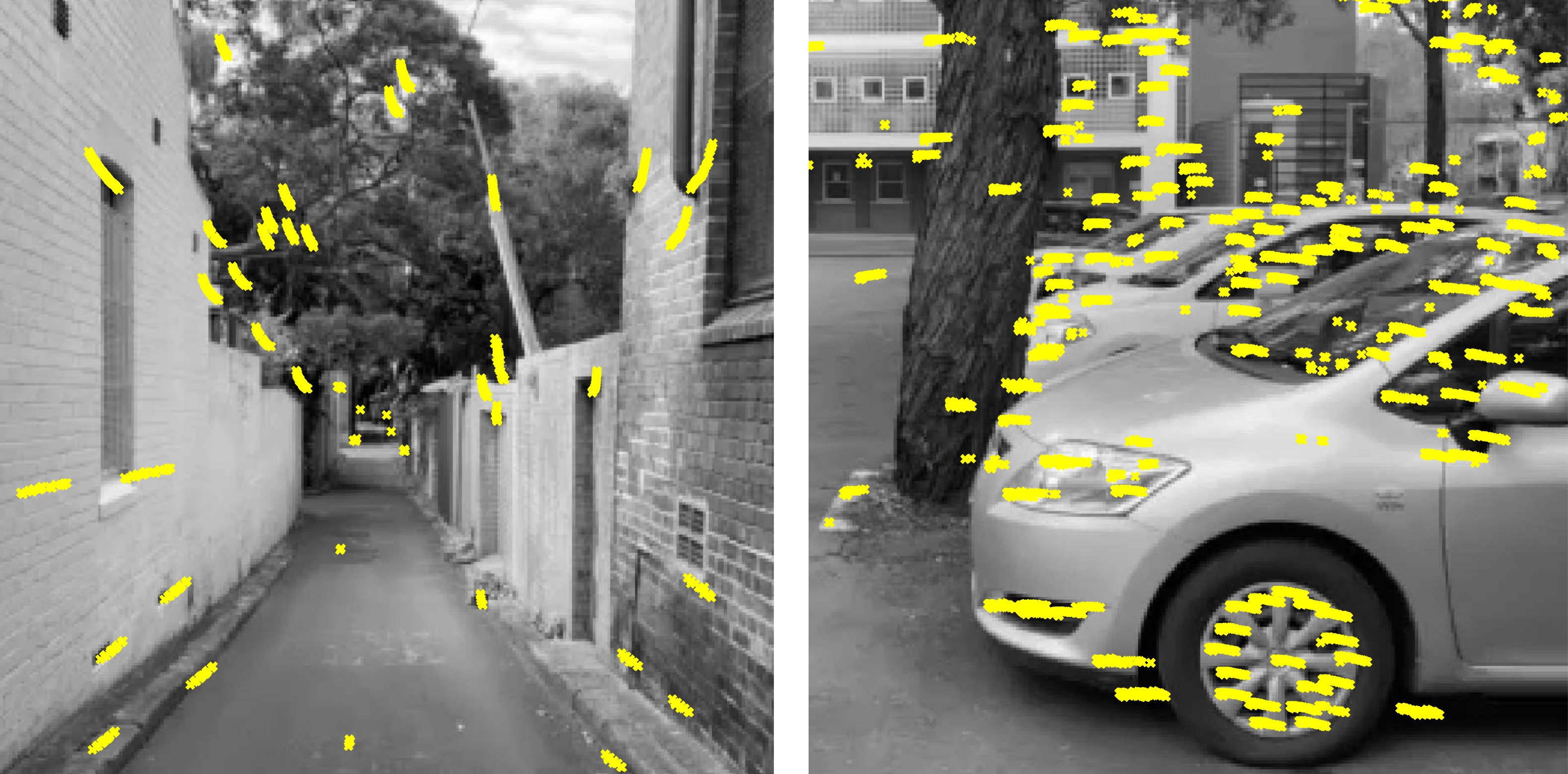}
	\caption{(left): In 3D scenes under general 6-\gls{DOF} platform motion, features in a burst exhibit apparent motion that is well approximated by line segments, even under platform rotation; (right): In the special case of a platform moving orthogonal to the principal axis of the camera, the apparent motion follows parallel line segments. We exploit these observations to search either 2D or 1D spaces of linear apparent motion to detect BuFF features.}
	\label{fig_localmotion}
\end{figure} 

%% Every descriptor represents an interesting point detected, and thus, allow matching algorithms to find correspondences among images. Most of the existing studies extending~\gls{SIFT} features enable the descriptors to extract spatio-temporal features~\cite{Al2012,Scovanner2007}. As low-light images are noise-limited, improving detector is important to reduce the spurious features and computation during reconstruction. In our work, we detect spatio-temporal features and describe each feature based on the histogram of edge orientations around its neighbourhood, as a 128-vector descriptor, similar to~\cite{Lowe2004}. As these descriptors are computed at the stage where images have higher~\gls{SNR} boost, they have better accuracy and selectivity advantage over conventional methods.

\section{Burst Feature Finder}
\label{sec:methods}

%%%%%%%%%%%%%%%% Methodology Chapter %%%%%%%%%%
\subsection{Apparent Motion within a Burst}
Capturing burst images using a camera mounted on a moving robotic platform exhibits fundamentally different motion profiles compared to hand-held burst photography. While handshake on handheld cameras contains significant high-frequency components, robotic platform dynamics are usually dominated by higher inertia and thus smoother instantaneous motion. This results in smooth and, for sufficiently fast bursts, locally linear apparent motion.

We identify two main variations in the local motion within a burst as shown in Fig.~\ref{fig_localmotion}. Features in a burst captured by a moving robot generally exhibit 2D linear apparent motion, especially when the robot is moving towards or away from the scene. On the other hand, when a robot is moving orthogonal to the principal axis of the camera, features within a burst exhibit 1D linear apparent motion. We build two variations of the feature finder,~\gls{BuFF} 2D for extracting features in a burst and~\gls{BuFF} 1D as a computational saving design to selectively operate on motion-constrained bursts. Further practical reasons for the design is described in~\ref{sec:methods_complexity}.

%%%%%%%%%%%%%%%% BFF Architecture from methods %%%%%%%%%%
\subsection{Motion Stack}
\label{sec:methods_motion}
\begin{figure}[t!] 
	\centering
	\includegraphics[width=0.85\columnwidth]{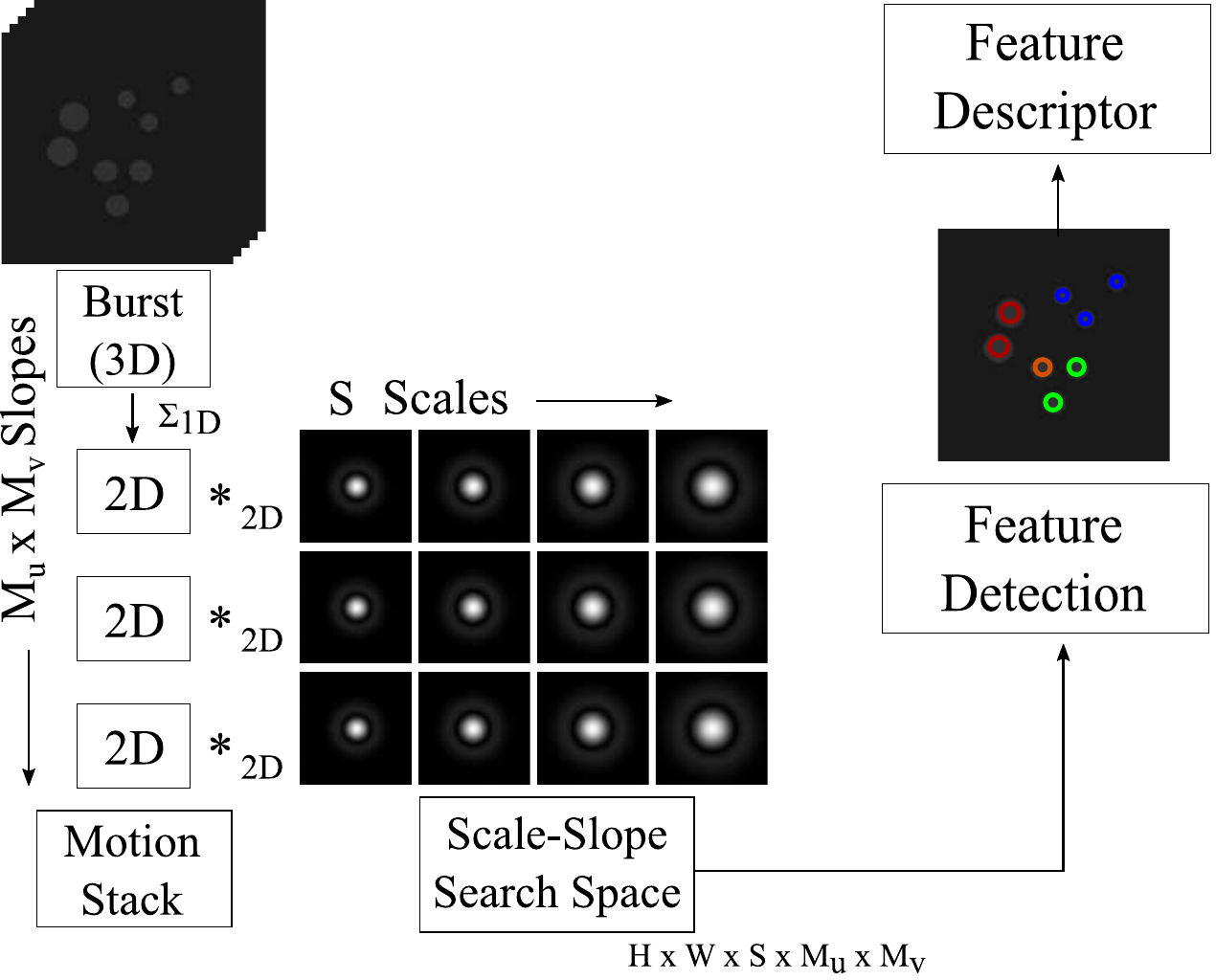}
	\caption{The proposed feature detection architecture: The $N$-frame input burst is converted to an $M_u \times M_v$ motion stack by shifting and summing frames across a range of putative apparent motions (``slopes''). This is reduced to an $M$-frame stack for 1D apparent motion. Each frame of the motion stack passes through a~\gls{DoG} scale-space filter, and features are detected as extrema in the resulting joint scale-slope search space. Each feature is described using a SIFT-style histogram of gradients applied to its corresponding motion-stack image, rather than the input frames. The resulting features have distinct location, scale and apparent motion, and exhibit high precision, recall, and matching performance.}
	\label{fig_bffconcept}
\end{figure}

\gls{BuFF} finds spatio-temporal features in a burst as in Fig.~\ref{fig_bffconcept}. We pass all the images in a burst $N$ through a shift-sum motion filter $H_M$. We enable 2D shift in the motion filter, by allowing two slope settings, $M_u$ corresponding to horizontal apparent motion and $M_v$ corresponding to vertical apparent motion of pixels.

We shift all the pixels within a burst by a given slope value that corresponds to the apparent motion of pixels and sum the motion-shifted images to compute a higher~\gls{SNR} image. We repeat this for a range of slope values $M$, as the apparent motion of the scene in foreground is generally higher than the background and have varying values depending on the scene. We build a motion stack with images, each representing a slope value. This is similar to a design of focal stack~\cite{Dansereau2015}, and allows depth selectivity.

%%%%%%%%%%%%%%%% BFF Architecture from methods %%%%%%%%%%
\subsection{Multi-Scale Multi-Motion Search Space}
We extend the search space of~\gls{SIFT} to a higher dimensional search by jointly searching for extrema across multiple scales and multiple slopes.

We convolve each image computed by the multi-motion stack with~\gls{DoG} filters $H_S$ over multiple scales $S$. We find extrema in a joint 5D search space $D_{5D}$ as in,
    \begin{equation}
    \label{eq:BuFF2D_searchspace}
    D_{5D}(\phi,\sigma,\lambda_{u,v}) = {{H_M(u,v,{\lambda_{u,v})}}*{H_S(u,v,\sigma)}}
    \end{equation}
where $u$ and $v$ are pixel positions in horizontal and vertical direction of the computed image from the motion stack respectively and $\phi$ = [$u$, $v$]. We apply the shift-sum motion $H_M$ filter over $M_u$ slopes in horizontal direction and $M_v$  slopes in vertical direction for all the images in the burst $N$ to build a 3D motion stack of $M_u$ x $M_v$ images. We find keypoints of distinct location $\phi$, scale $\sigma$ and apparent motion $\lambda$. This rejects spurious features from the noisy images. 

BuFF searches through location, scale, and a 2D apparent motion space to build a 5D search space. In the special case of 1D apparent motion, this reduces to a 4D search space.

While~\gls{BuFF} has similar parameters to~\gls{SIFT} including octaves, levels, contrast threshold and edge threshold, the key variable in~\gls{BuFF} is the range of slopes. As~\gls{BuFF} finds features in a burst, number of images in a burst is also a variable. Adding more images improves the performance asymptotically. 

\subsection{Descriptor}
Similar to~\gls{SIFT} and~\gls{LiFF}, we compute a histogram of edge orientations for each features, a 128-element vector. In addition to the sensitivity and selectivity of the detection, describing features at a stage where the images have higher~\gls{SNR} compared to the input images in the burst allow robust reconstruction of light-constrained environment. Because we use the images computed from the motion stack to build the search space, we also have inherent depth information, that pave way for other applications including segmentation and depth selection.

\subsection{Complexity}
\label{sec:methods_complexity}
Choosing the appropriate order of events to design the search-scale space yields computational savings.~\gls{BuFF} 1D is carried in the order of building a multi-motion stack of images from burst images, and then convolving the shifted images with~\gls{DoG} filters. Because the convolution operation is more computational expensive than shifting and summing pixels, the order of building multi-motion stack prior to multi-scale stack is efficient for~\gls{BuFF} 1D.

As~\gls{BuFF} 2D have larger search space comparing to~\gls{BuFF} 1D, designing multi-motion stack prior to the convolution is not desirable. Instead, convolving burst images with~\gls{DoG} filters and building multi-motion stack using images from multi-scale stack is appropriate for computation.

\section{Results}
\label{sec:results}

\begin{figure*}[t!]
	\centering
	\includegraphics[width=0.98\textwidth]{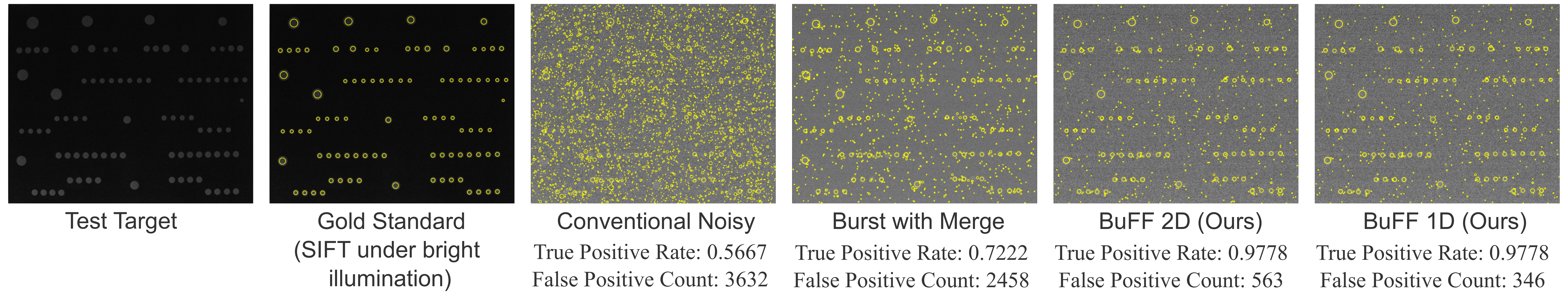}
	\caption{Validation with a printed test target: (left to right) Test target with disks of varying scales to demonstrate feature performance. SIFT finds features well under bright illumination, but suffers from many false positives and reduced true positives in low light. Employing conventional burst imaging prior to applying SIFT improves performance, but BuFF 2D and 1D show much greater performance due to the joint scale-motion search -- see also Fig.~\ref{fig_roc}.}
	\label{fig_features}
\end{figure*}

\begin{figure}[ht]
	\centering
	\includegraphics[width=0.9\columnwidth]{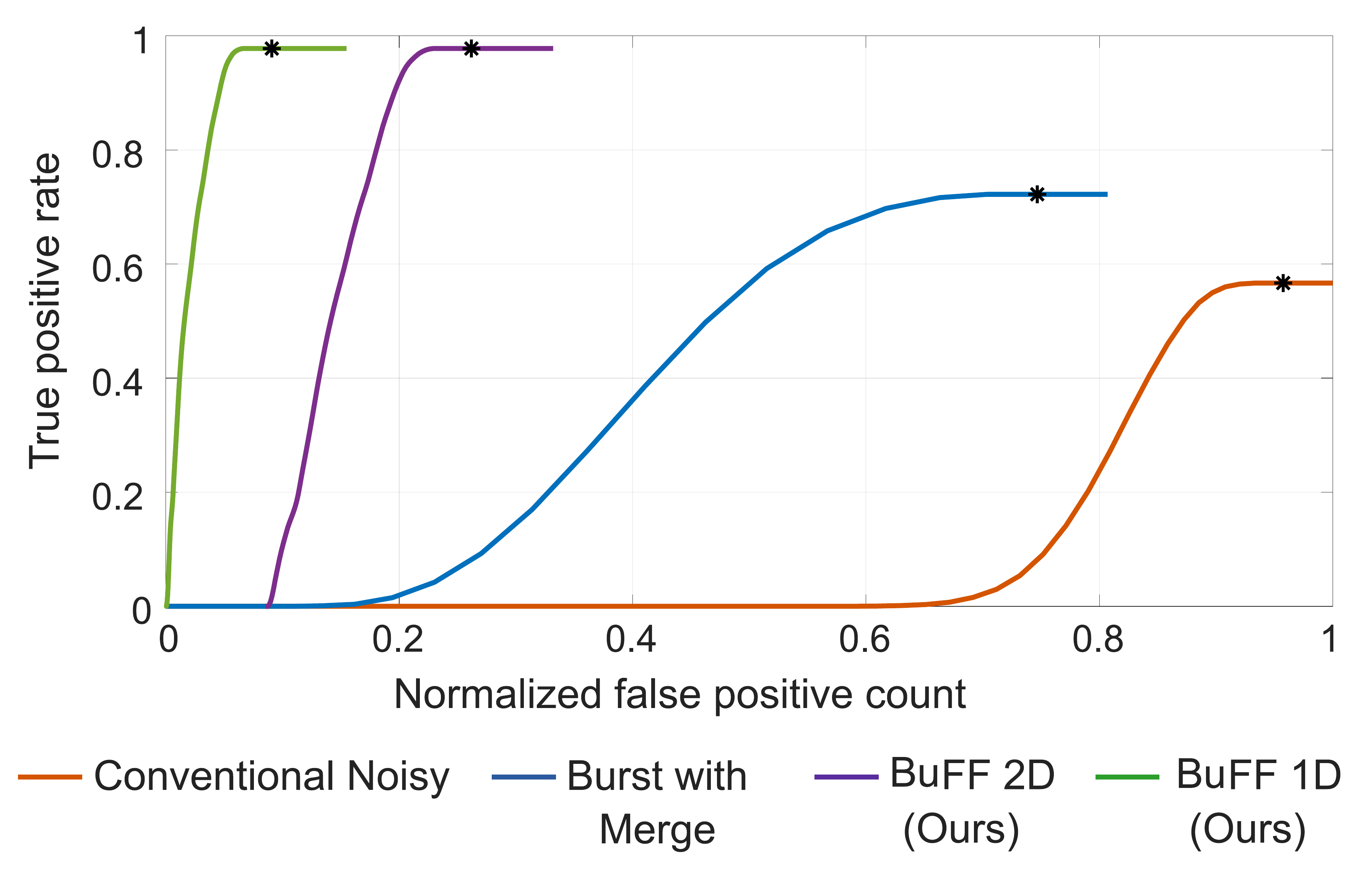}
	\caption{ROC curves for (orange) \gls{SIFT} on conventional imagery, (blue) SIFT on merged burst imagery, (violet) \gls{BuFF} with 2D linear local apparent motion, and  (green) \gls{BuFF} with 1D linear local apparent motion. BuFF 1D admits fewer spurious features because of the smaller search space, and both variants of BuFF exhibit higher true positive rates because of the signal boost associated with joint scale-motion search. We use the ROC curves to select comparable peak thresholds for each method, such that 10\% of detected features are false positives.}
	\label{fig_roc}
\end{figure}
%%%%%%%%%%%%%%%% Proof-of-Concept %%%%%%%%%%
In the following, we first evaluate the performance of our feature detector and descriptor in noisy imagery using captured burst of a target scene. We initially use a target scene with varying scales to analyse the nature of feature detection. Then, we quantitatively evaluate the feature performance of \gls{BuFF} in an SfM pipeline across various scenes, by comparing against \gls{SIFT} on conventional noisy images and \gls{SIFT} on burst-merged images by considering reconstruction performance and pose estimation.

\subsection{Feature Performance in Noise}
\label{sec:results_feature}
We print a test target of 90 disks at varying scales with known feature locations to demonstrate feature performance (see Fig. \ref{fig_features}, left). The color contrast ratio between the disks and the background is 1.2.

We capture bursts of images using a Basler 1600-60um monocular machine vision camera with an f/11 lens. We capture 12-bit monocrhome images of size 1600x1200. The~\gls{EV} is selected to match an f/2 camera with 0.16ms exposure time and 81.18 lux of illumination, or equivalently a 5ms exposure with 1.18 lux of illumination. We also capture the test target scene over 100ms exposure time, yielding a high~\gls{SNR} image which we use as a gold standard.  

We examine the true positive rate and the false positive count of the captured imagery of the target scene for a sweeping peak threshold as shown in Fig.~\ref{fig_roc}. We compare our methods against the~\gls{ROC} of alternative approaches: the conventional noisy image and burst-merged image. We select the peak threshold at which, each methods perform preferably at its best for the input noise level of the imagery, i.e. selecting a peak threshold value at the highest true positive range with an additional 10\% of total false positive features. Extracted features using alternative approaches and our methods on the associated imagery at the selected peak threshold are shown in Fig.~\ref{fig_features}. 

Conventional methods find fewer true positive features and more spurious features for the captured imagery. Burst-merged images find more true positive features and fewer spurious features comparing to conventional methods. Our methods outperform both conventional noisy and burst-merged image finding higher overall true positive features. As the search space is larger for~\gls{BuFF} 2D comparing to~\gls{BuFF} 1D, between our methods,~\gls{BuFF} 1D finds fewer spurious features comparing to~\gls{BuFF} 2D. 
\begin{figure}[ht!]
	\centering
	\includegraphics[width=\columnwidth]{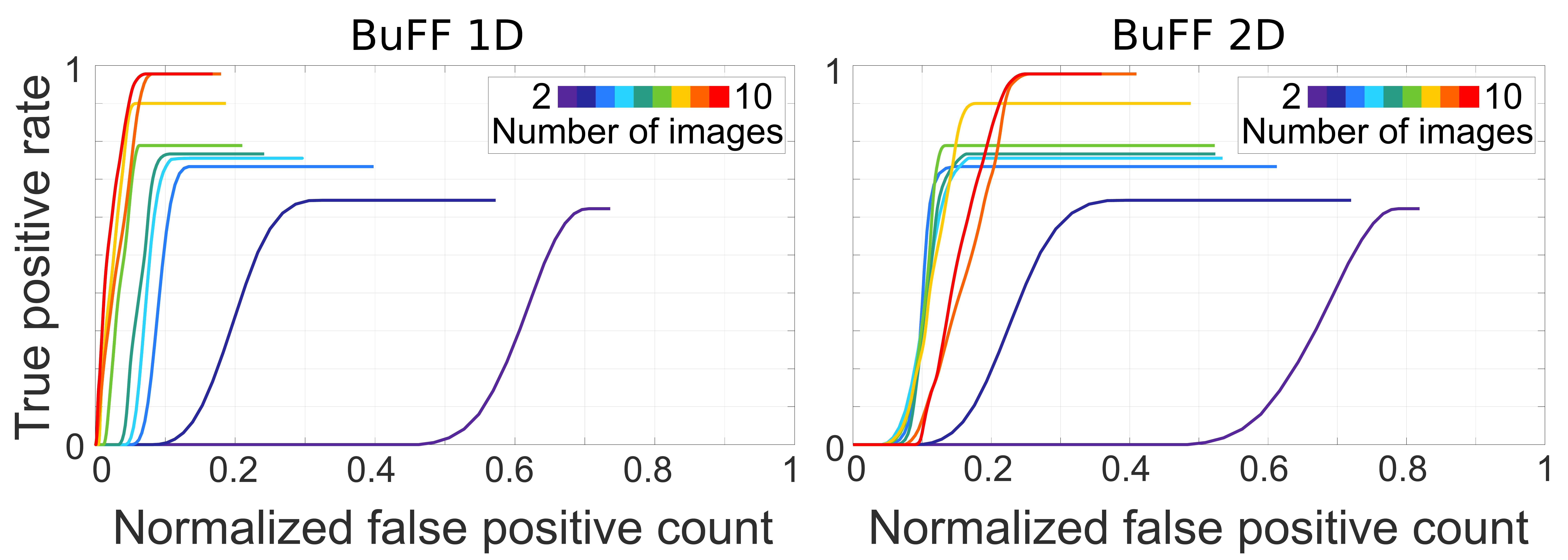}
	\caption{Impact of number of frames in a burst: employing more images improves both spurious feature count and true positive rate. These performance curves vary with camera, scene, and platform motion characteristics.  For this  scenario, performance saturates at around 10 images per burst.} 
	\label{fig_performance}
\end{figure}

We additionally compare the variation in feature performance of our methods for different number of images in a burst as shown in Fig.~\ref{fig_performance}. As we increase the number of images in a burst, the~\gls{SNR} of the images used in constructing the multi-scale multi-motion search space increases. This gives an advantage for both the detector and the descriptor to find higher true features and fewer spurious features. After six images, there is minimal difference in extracted true features and spurious features for the noise level of the captured imagery.

\begin{figure}[b!]
	\centering
	\includegraphics[width=0.8\columnwidth]{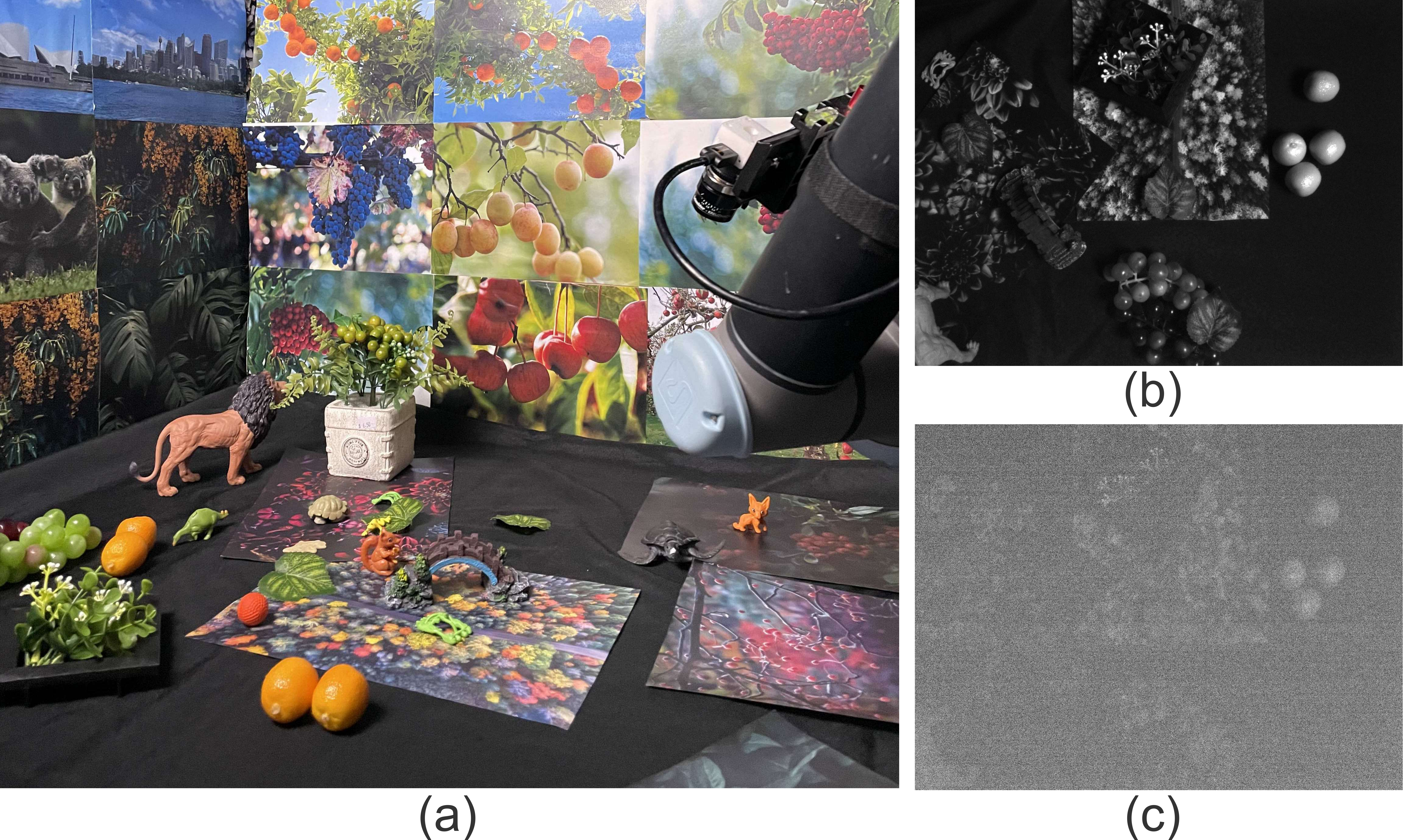}
	\caption{Examples of captured imagery: (a)~View of a well-lit scene and arm-mounted camera; (b)~a gold standard image captured over 100ms exposure time; and (c) a single noisy image captured over 5ms at the same camera pose. The full dataset contains 10 trajectories, each with 20 bursts of 10 images.}
	\label{fig_experiment}
\end{figure}

%%%%%%%%%%%%%%%% Reconstruction %%%%%%%%%%%%%%%%%%%%%%
\subsection{Reconstruction Accuracy}
\label{sec:results_reconstruction}

%%%%%%%%%%%%%%%%%%%%%%%% Table 1 Placement %%%%%%%%%%%%%%%%%%%%%%%%%%%%%%%%%%%%%%%%%

\begin{table*}
\centering
\caption{Average performance of 3D reconstruction for light-constrained scenes with 1D and 2D apparent motion}
\label{tab:features}
\begin{tabular}{clrrrrrrrrr} 
\hline
\begin{tabular}[c]{@{}c@{}}Apparent\\Motion\end{tabular} & \multicolumn{1}{c}{Method} & \multicolumn{1}{c}{\begin{tabular}[c]{@{}c@{}}Convergence\\ Rate\end{tabular}} & \multicolumn{1}{c}{\begin{tabular}[c]{@{}c@{}}Images\\ Pass \%\end{tabular}} & \multicolumn{1}{c}{\begin{tabular}[c]{@{}c@{}}Keypoints/\\ Image\end{tabular}} & \multicolumn{1}{c}{\begin{tabular}[c]{@{}c@{}}Putative\\ Matches/\\ Image\end{tabular}} & \multicolumn{1}{c}{\begin{tabular}[c]{@{}c@{}}Inliers\\ Matches/\\ Image\end{tabular}} & \multicolumn{1}{c}{\begin{tabular}[c]{@{}c@{}}Match\\ Ratio\end{tabular}} & \multicolumn{1}{c}{\begin{tabular}[c]{@{}c@{}}Match\\ Score\end{tabular}} & \multicolumn{1}{c}{Precision} & \multicolumn{1}{c}{\begin{tabular}[c]{@{}c@{}}3D Points/\\ Image\end{tabular}} \\ 
\hline
\multirow{5}{*}{1D} & Gold Standard & 1.0 & 100 & 2881 & 1.175x10\textsuperscript{4} & 1.170x10\textsuperscript{4} & 4.161 & 4.141 & 0.995 & 5159 \\
 & Conventional Noisy & 0.0 & 0 & 5574 & 68.37 & 64.11 & 0.012 & 0.012 & 0.873 & - \\
 & Burst with Merge & 0.8 & 59 & 2216 & 361.8 & 332.3 & 0.185 & 0.170 & 0.934 & 457.2 \\
 & BuFF 1D & ~\color[HTML]{34A853}\textbf{1.0} & ~\color[HTML]{34A853}\textbf{98} & 1386 & ~\color[HTML]{34A853}\textbf{384.1} & ~\color[HTML]{34A853}\textbf{374.6} & ~\color[HTML]{34A853}\textbf{0.280} & ~\color[HTML]{34A853}\textbf{0.273} & 0.976 & ~\color[HTML]{34A853}\textbf{521.2} \\
 & BuFF 2D & ~\color[HTML]{34A853}\textbf{1.0} & 92 & 1636 & 345.5 & 338.4 & 0.230 & 0.225 & ~\color[HTML]{34A853}\textbf{0.980} & 490.4 \\ 
\hline
\multirow{4}{*}{2D} & Gold Standard & ~1.0 & 100 & 3270 & 1.303x10\textsuperscript{4} & 1.296x10\textsuperscript{4} & 3.893 & 3.873 & 0.994 & 6110 \\
 & Conventional Noisy & 0.0 & 0 & 5680 & 18.41 & 14.25 & 0.003 & 0.003 & 0.860 & - \\
 & Burst with Merge & 0.8 & 77 & 2746 & 485.7 & 457.0 & 0.177 & 0.167 & 0.954 & 617.8 \\
 & BuFF 2D & ~\color[HTML]{34A853}\textbf{1.0} & \color[HTML]{34A853}\textbf{98} & 2144 & \color[HTML]{34A853}\textbf{565.6} & \color[HTML]{34A853}\textbf{551.4} & \color[HTML]{34A853}\textbf{0.263} & \color[HTML]{34A853}\textbf{0.257} & \color[HTML]{34A853}\textbf{0.978} & \color[HTML]{34A853}\textbf{732.0} \\
\hline
\end{tabular}
\end{table*}

We demonstrate our method for the reconstruction experiments by mounting the same camera under the same lighting conditions as discussed in Sec.~\ref{sec:results_feature} on a UR5e robotic arm as shown in Fig.~\ref{fig_experiment}. We capture 20 burst sequences with 10 images in each burst for five trajectories. We capture these sequences while the robot is moving towards and away from the scene, resulting in burst sequences having approximately 2D apparent motion between frames. We also capture bursts having 1D apparent motion between frames in a similar fashion, by capturing multiple burst sequences for five trajectories while the robot is moving orthogonal to the principal axis of the camera.

The scene is designed similar to a forest environment with objects with different textures, shapes and sizes under controlled lighting as shown in Fig.~\ref{fig_experiment}. We capture our dataset over 5ms and 100ms exposure times as noisy and gold standard images respectively to correspond closely to the noise levels of the captured test target scene. We also capture bias frames at the same exposure and gain to remove fixed-pattern noise. The apparent motion within each burst varies with a maximum value of 30 pixels.

We compute features from the variants of \gls{BuFF} and the VLFeat\footnote[1]{https://www.vlfeat.org/overview/sift.html} implementation of SIFT using selected peak threshold as explained in Fig. \ref{fig_roc}, edge threshold 10, and \gls{DoG} scales covering 6 octaves over 4 levels per octave. This is similar to the settings used on gold standard test target scene to generate all the true features with no spurious features. For the variants of \gls{BuFF}, we computed motion stack over 7 slopes in horizontal and vertical directions accordingly. For the descriptor, we apply L1 root normalization across all methods, similar to \cite{Dansereau2019} to yield improved matches.

Following the feature comparison approach in~\cite{Schonberger2017}, we evaluate reconstruction performance of motion constrained trajectories captured as multiple bursts with 1D apparent motion as shown in \Table{\ref{tab:features}} in terms of convergence rate: number of reconstructed scenes refined by bundle adjustment given all scenes, images pass: number of images used for reconstruction given the input images, number of keypoints per image, putative feature matches per image, inliers per image, match ratio: the number of detected features classified as putative matches, match score: the number of detected features yielding inlier matches, precision: the proportion of putative matches, and the average 3D points in the reconstructed models per image. In \Table{\ref{tab:features}}, we highlight the best results in green, and they correspond to the two variants of our methods. We compare~\gls{BuFF} 2D for trajectories with 1D motion, and while it uses 92\% of all images to reconstruct light-constrained scenes with competitive putative matches, inlier matches and 3D points per image, 1D subset aligned in the direction of the apparent motion demonstrate more matches and 3D points overall. This is because of less spurious feature detection in limited search space comparing to~\gls{BuFF} 2D. 

Evaluating trajectories with bursts having 2D apparent motion between frames, our method outperforms alternative approaches across all metrics, by reconstructing 98\% of all the input scenes. Conventional single noisy images failed to identify a good initial image pair and failed to converge during bundle adjustment. Our method reconstructs hundreds more matches and 3D points per image compared with burst-merged reconstruction. It also shows higher match ratio, match score and precision for the extracted features.

\begin{table*}[ht!]
\centering
\caption{Mean translation error and mean rotational error in camera poses for scenes with 1D and 2D apparent motion}
\label{tab:camera}
\begin{tabular}{llrrrr} 
\hline
\multirow{2}{*}{\begin{tabular}[c]{@{}l@{}}Apparent\\motion\end{tabular}} & \multirow{2}{*}{Method} & \multicolumn{2}{c}{Absolute trajectory error} & \multicolumn{2}{c}{Relative pose error} \\
 &  & \multicolumn{1}{c}{trans.(cm)} & \multicolumn{1}{c}{rot.(deg)} & \multicolumn{1}{c}{trans.(cm)} & \multicolumn{1}{c}{rot.(deg)} \\ 
\hline
\multirow{3}{*}{1D} & Burst with Merge & 2.84 & 2.07 & 4.97 & 0.73 \\
 & BuFF 1D & \color[HTML]{34A853} ~\textbf{1.23} & 1.24 & \color[HTML]{34A853} ~\textbf{2.00} & 0.05 \\
 & BuFF 2D & 1.27 & \color[HTML]{34A853} ~\textbf{0.62} & 2.01 & \color[HTML]{34A853} ~\textbf{0.02} \\ 
\hline
\multirow{2}{*}{2D} & Burst with Merge & 1.80 & 1.11 & 3.60 & 0.04 \\
 & BuFF 2D & \color[HTML]{34A853}~\textbf{1.44} & \color[HTML]{34A853} ~\textbf{0.58} & \color[HTML]{34A853} ~\textbf{2.55} & \color[HTML]{34A853} ~\textbf{0.03} \\ 
\hline
\end{tabular}
\end{table*}

%%%%%%%%%%%%%%%% Camera Trajectory Figure Placement %%%%%%%%%%
\begin{figure}[t!]
	\centering
	\includegraphics[width=\columnwidth]{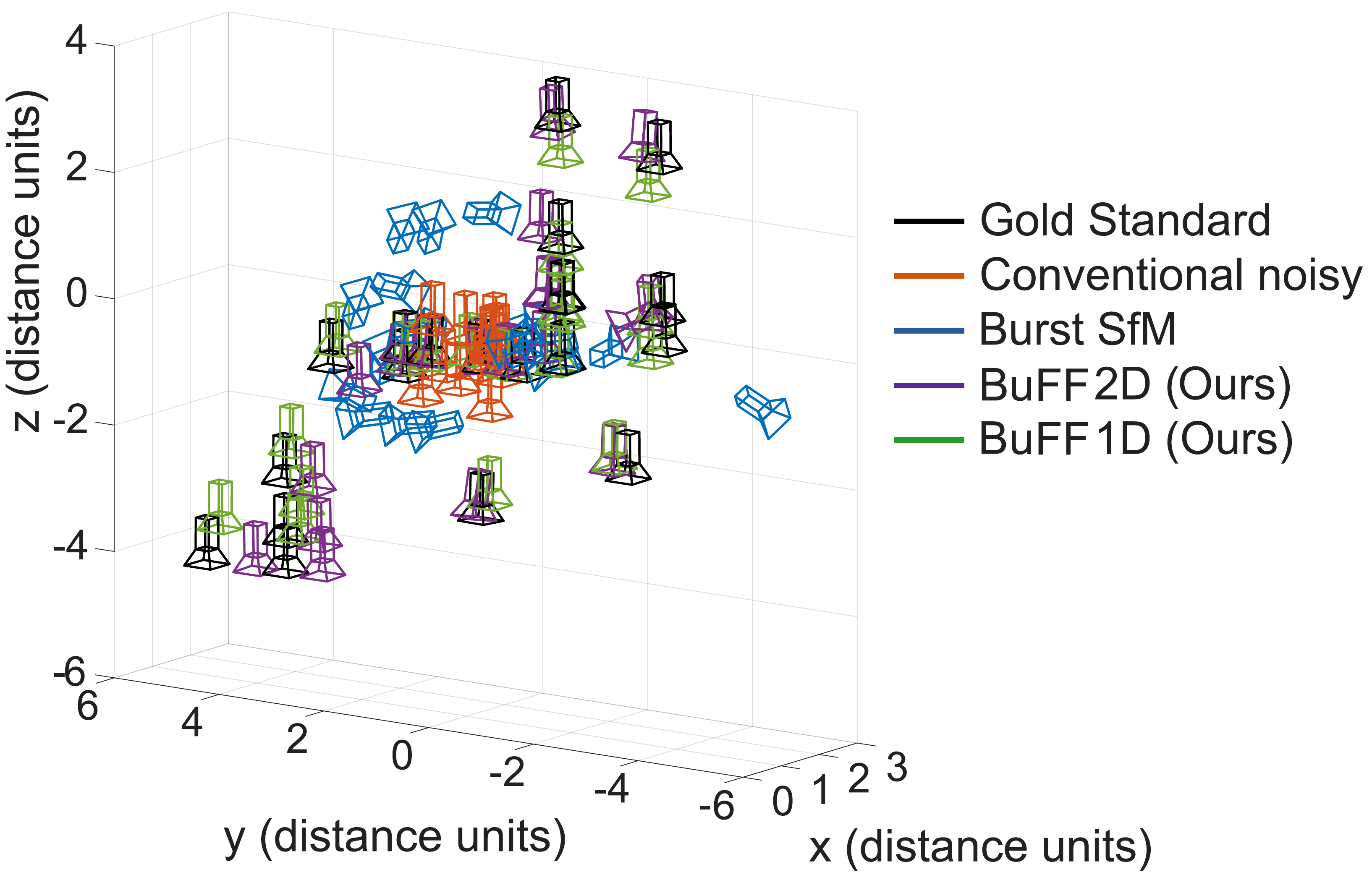}
	\caption{Localisation performance: Conventional imagery yields very poor camera pose estimates (orange) and even fails to converge without relaxing inlier rate settings. Employing burst-merged images reconstructed 90\% of input images but still showed poor pose estimation (blue). The proposed \gls{BuFF} variants successfully employed the most images and yielded the most accurate pose estimates, with the 2D variant (violet) showing slightly lower accuracy than BuFF 1D (green), due to increased spurious features associated with the larger 2D search space. See Table~\ref{tab:camera} for quantitative results.}
	\label{fig_camera}
\end{figure}

%%%%%%%%%%%%%%%% Camera Trajectory %%%%%%%%%%
\subsection{Pose Estimation}
We compute average absolute trajectory error and relative pose error for translation and rotation from the poses estimated by COLMAP as shown in~\Table{\ref{tab:camera}}. The color scheme matches that used in~\Table{\ref{tab:features}}. The results show the two \gls{BuFF} variants outperforming burst with merge in all metrics with \gls{BuFF} 2D showing a slight advantage over \gls{BuFF} 1D for rotation estimates. Because conventional noisy method fails to converge throughout, there are no camera pose estimates for that method. However, for qualitative comparison as shown in Fig.~\ref{fig_camera}, we relax the reconstruction settings for a scene by reducing the inliers matches required to select an initial pair. 

We align and scale the camera poses to the gold standard poses as there is an arbitrary scale factor involved in monocular~\gls{SfM}. The arbitrary scale used in describing the poses are determined by the distance between the first pair of registered images. While relaxing inlier rate settings encourage conventional method to converge, the camera pose estimates are inaccurate as shown in Fig.~\ref{fig_camera}. Estimating camera pose estimates using burst-merged images show poor estimates and does not reconstruct all camera poses. Both variants of our method compute accurate camera pose estimates comparing to the competing methods. 

%%%%%%%%%%%%%%%% Speed Discussion %%%%%%%%%%

\subsection{Speed}
For a dataset of 200 images in 20 bursts, our MATLAB implementation for detecting and describing burst features and reconstruction using imported features of BuFF 1D took 18.42 minutes and BuFF 2D took 108.5 minutes on an Intel i7-9700 at 4.70 GHz as in~\Table{\ref{tab:speed}}. This reported time for \gls{BuFF} 2D is by following the design of building motion stack prior to convolution stage with scale stack. We expect this could be accelerated substantially. By contrast, operating~\gls{SIFT} MATLAB implementation on 20 conventional noisy images took 15.04 minutes, and operating on burst-merged images with align and merge took 13.86 minutes.

\begin{table}[h]
\centering
\caption{Time taken for reconstruction including feature extraction}
\label{tab:speed}
\begin{tabular}{ll}
Method & \begin{tabular}[c]{@{}l@{}}Time taken\\(min)\end{tabular}  \\ \hline
Conventional Noisy & 15.04 \\
Burst with Merge & 13.86 \\
BuFF 2D (Ours) & 108.5 \\
BuFF 1D (Ours) & 18.42 \\ \hline
\end{tabular}
\end{table}

%%%%%%%%%%%%%%%%%% Limitation Discussion and Intuitions %%%%%%%%%%%%%%%%%%%%%%%%%%
\subsection{Failure cases}
\begin{figure}[h]
	\centering
	\includegraphics[width=\columnwidth]{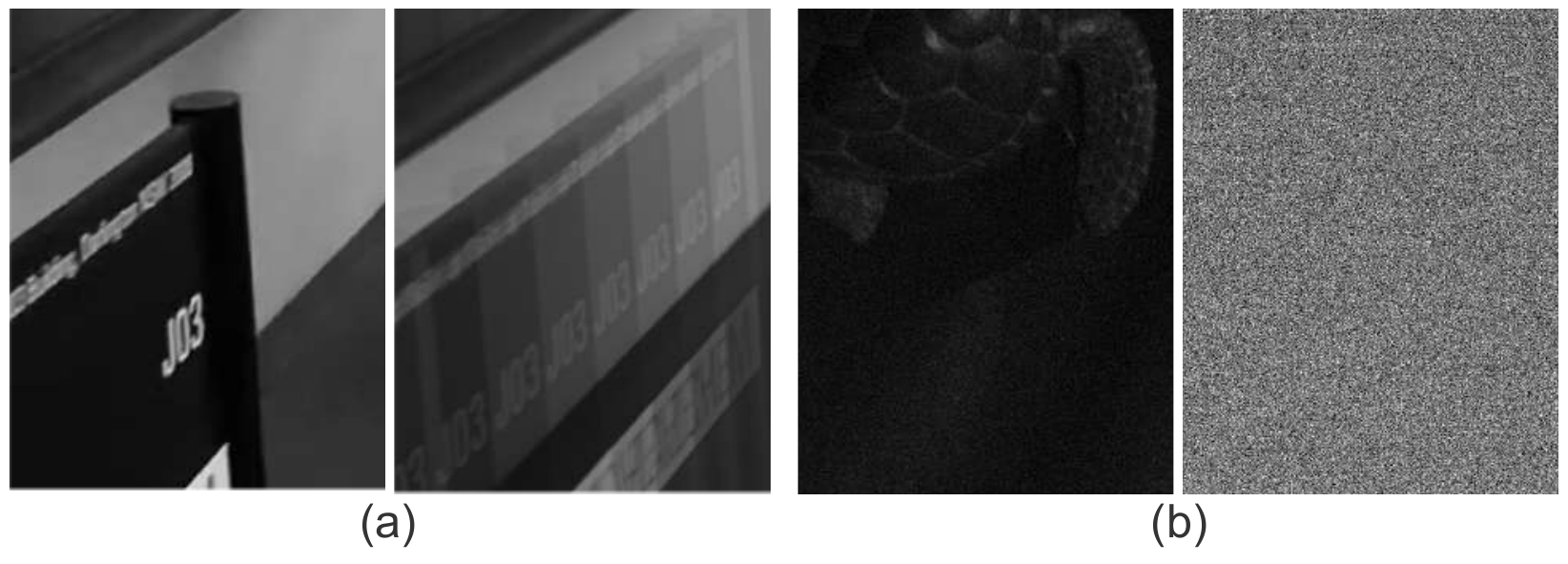}
	\caption{Failure cases: (a, left) A single input frame and (right) an image from the motion stack with large putative apparent motion showing false structure due to aliasing. Employing faster bursts and reducing putative apparent motion eliminates this kind of aliasing. (b, left) A low-contrast textured scene in bright illumination, and (right) the same under low light showing insufficient signal for feature detection. Ultimately there needs to be enough signal for BuFF to find features.}
	\label{fig_failedcase}
\end{figure}

Examples of failed cases with larger apparent motion between frames and extremely low contrast are shown in Fig.~\ref{fig_failedcase}. Capturing faster bursts eliminates the aliasing effect. The approach is also affected by the amount of signal in the original image and we believe capturing a burst with more images will benefit feature detection. 

\section{Conclusions}
\label{sec:concl}

We introduced a novel burst feature finder to reconstruct low light scenes. We captured multiple-bursts sequences along typical motion trajectories and detected spatio-temporal features in a multi-scale multi-motion search space. We employed a descriptor that provided additional advantage with higher~\gls{SNR} boost comparing to conventional methods.

We demonstrated improved performance comparing to conventional imaging and burst-merged reconstruction in terms of true features, spurious features, convergence rate, match score, match ratio, precision and 3D points per images. We show that finding features directly in a burst estimates accurate camera poses comparing to finding features on burst-merged images or conventional images for low light scenes. We expect that this work can improve 3D reconstruction in low-light, allowing robots to perform better in challenging rescue missions and low light delivery.

This work reformulates the trajectory of a robot as multiple-bursts, and find burst features in low light to improve 3D reconstruction. For future work, we anticipate learning-based approaches for motion stack design, deep burst feature extraction and object tracking of visually challenging features like snow. We also expect complementary sensors to aid in improving reconstruction in low light.
%%%%%%%%%%%%%%%%%%%%%%%%%%%%%%%%%%%%%%%%%%%%%%%%%%%%%%%%%%%%%%%%%%%%%%%%%%%%%%%%

%\section*{APPENDIX}
%
%Appendixes should appear before the acknowledgment.
%%%%%%%%%%%%%%%%%%%%%%%%%%%%%%%%%%%%%%%%%%%%%%%%%%%%%%%%%%%%%%%%%%%%%%%%%%%%%%%%

%\begin{thebibliography}{99}
{\small
	\bibliographystyle{IEEEtran}
	\bibliography{./references}
}
%\end{thebibliography}
\end{document}